\documentclass[fleqn,10pt]{wlscirep}
\usepackage[utf8]{inputenc}
\usepackage[T1]{fontenc}
\usepackage{lineno}
\usepackage{multirow}
\usepackage{makecell}
\usepackage{float}

\title{DDTracking: A Deep Generative Framework for Diffusion MRI Tractography with Streamline Local-Global Spatiotemporal Modeling}

\author[1]{Yijie Li}
\author[1]{Wei Zhang}
\author[1]{Xi Zhu}
\author[2]{Ye Wu}
\author[3]{Yogesh Rathi}
\author[3,*]{Lauren J. O’Donnell}
\author[1,*]{Fan Zhang}
\affil[1]{University of Electronic Science and Technology of China, Chengdu, China}
\affil[2]{Nanjing University of Science and Technology, Nanjing, China}
\affil[3]{Brigham and Women’s Hospital, Harvard Medical School, Boston, USA}

\affil[*]{Lauren J O’Donnell (odonnell@bwh.harvard.edu) and Fan Zhang (fan.zhang@uestc.edu.cn) are co-senior-authors.}

\begin{abstract}
This paper presents DDTracking, a novel deep generative framework for diffusion MRI (dMRI) tractography that formulates streamline propagation as a conditional denoising diffusion process. In DDTracking, we introduce a dual-pathway encoding network that jointly models local spatial encoding (capturing fine-scale structural details at each streamline point) and global temporal dependencies (ensuring long-range consistency across the entire streamline). Furthermore, we design a conditional diffusion model module, which leverages the learned local and global embeddings to predict streamline propagation orientations for tractography in an end-to-end trainable manner. We conduct a comprehensive evaluation across diverse, independently acquired dMRI datasets, including both synthetic and clinical data. Experiments on two well-established benchmarks with ground truth (ISMRM Challenge and TractoInferno) demonstrate that DDTracking largely outperforms current state-of-the-art tractography methods. Furthermore, our results highlight DDTracking’s strong generalizability across heterogeneous datasets, spanning varying health conditions, age groups, imaging protocols, and scanner types. Collectively, DDTracking offers anatomically plausible and robust tractography, presenting a scalable, adaptable, and end-to-end learnable solution for broad dMRI applications. Code is available at: \underline{https://github.com/yishengpoxiao/DDtracking.git}.
\end{abstract}
\begin{document}

\flushbottom
\maketitle

\thispagestyle{empty}

\section{Introduction}
\label{sec:introduction}

Diffusion MRI (dMRI) tractography is an advanced imaging technique for in vivo mapping of white matter (WM) pathways and structural connectivity in the brain at the macro scale \cite{basser1994mr,basser2000vivo,zhang2022quantitative}. This methodology capitalizes on the anisotropic diffusion of water molecules, which is modulated by the microstructural features of brain tissue, particularly axonal membranes \cite{o2011introduction}. By modeling these diffusion patterns, tractography infers virtual streamlines representing macroscopic WM pathways, thus providing critical insights into the brain’s functional organization and connectivity.

Traditional model-based tractography methods convert dMRI signals into virtual streamlines, iteratively propagated from predefined seed points using either deterministic or probabilistic approaches \cite{malcolm2010filtered,reddy2016joint,aganj2009odf,tournier2012mrtrix,leemans2025deterministic,girard2025probabilistic}. Deterministic algorithms utilize rigid orientation criteria, ensuring computational efficiency but struggling to reconstruct complex fiber structures accurately, such as crossing or branching fibers. On the other hand, probabilistic methods employ Monte Carlo sampling of fiber orientation distributions, providing greater flexibility and precision, albeit at the cost of increased computational demands. Both frameworks rely on the evolution of physical models, from the single Gaussian diffusion assumption in diffusion tensor imaging (DTI) to advanced models like multi-compartment models, fiber orientation distribution (FOD), and Q-ball imaging \cite{basser1994mr,descoteaux2007regularized,tournier2004direct,jeurissen2014multi}, which better represent multiple fiber orientations within a voxel. Recent improvements have incorporated external constraints, such as anatomically constrained tractography guided by structural MRI to delineate gray-white matter boundaries \cite{smith2012anatomically} and microstructure-informed tractography leveraging axonal diameter estimates from techniques like AxCaliber, further enhancing the precision of fiber reconstruction \cite{daducci2016microstructure}. Despite these advances, model-based methods remain sensitive to noise and artifacts and are prone to accumulating local orientation biases that can distort streamline trajectories over long distances. Additionally, the reliance on complex modeling and tracking algorithms imposes significant computational demands and heightens sensitivity to data quality.

The advent of machine learning (ML) has introduced promising solutions to these longstanding challenges \cite{poulin2019tractography,karimi2024diffusion,zhang2025think}. Unlike traditional model-based methods, ML approaches are data-driven and do not rely on predefined physical models, which can introduce simplifications and bias. By integrating diverse information sources, ML tractography enables enhanced, accurate, and robust maps of WM pathways. The pioneering work by Neher et al. introduced the first ML method \cite{neher2017fiber}, which employs a Random Forest classifier for streamline propagation, and demonstrated the feasibility of using ML to learn tracking decisions from the diffusion signals and streamline trajectories. Subsequent efforts utilizing recurrent neural networks (RNNs) and gated recurrent units (GRUs) have further enhanced the ability to model long-term dependencies in streamline orientation \cite{cai2024convolutional,poulin2017learn,benou2019deeptract}, addressing issues related to the persistence of local orientation biases in traditional tractography approaches. More recently, methods such as Learn to Track \cite{poulin2017learn} and DeepTract \cite{benou2019deeptract} have shown the potential to address the challenges of fiber crossing and smooth streamline propagation by incorporating information from both local and previous streamline steps. Contemporary probabilistic frameworks, such as Entrack, leverage multilayer perceptrons (MLPs) to quantify uncertainty in streamline predictions using Fisher-von-Mises (FvM) distributions \cite{wegmayr2021entrack}. Track-to-Learn, a deep reinforcement learning-based tractography method \cite{theberge2021track,theberge2024matters}, utilizes a reward function to optimize streamline tracing based on local information, promoting closer alignment with major FOD peaks while ensuring smooth propagation. However, these methods typically rely on approximations of fiber orientation distributions or discrete orientation sampling strategies, which may fail to capture the complex variations in fiber orientation. Additionally, the inability to directly predict streamline orientation at each location can lead to potentially anatomically implausible streamlines.

Diffusion models are gaining popularity in applications such as object tracking and motion planning \cite{janner2022planning,gu2022stochastic,lv2024diffmot,luo2024diffusiontrack}, tasks that share conceptual parallels with tractography due to their capacity to model complex trajectory distributions. In general, diffusion models operate via a two-stage process \cite{ho2020denoising}: a forward diffusion stage, where input data is progressively perturbed by Gaussian noise, and a reverse diffusion stage, which aims to gradually denoise and recover the original input. This framework has demonstrated remarkable flexibility and adaptability in the computer vision field, particularly in modeling complex, multimodal trajectory distributions. For example, popularly used diffusion models like Diffuser \cite{janner2022planning} and DiffMOT \cite{lv2024diffmot} generate goal-conditioned trajectories of agents and tracked objects through iterative noise refinement, enabling flexible and accurate motion prediction. Similar to these tasks in computer vision, an essential capability for fiber tracking is to infer a streamline trajectory from dMRI signals. Thus, diffusion models have strong potential to model the complex distribution of streamline trajectories in the brain. In our study, we propose to formulate fiber tracking as a goal-conditioned sequence generation task, where the objective is to generate anatomically plausible streamline propagation orientations conditioned on the dMRI signals. By training diffusion models on high-confidence tractography data, we aim to enable them to learn and represent the complex fiber configurations, including crossing, fanning, and branching patterns that frequently challenge conventional deterministic or probabilistic methods.

In our study, we propose a novel deep diffusion model framework, \textit{DDTracking}, to estimate white matter streamline propagation orientation for tractography. Our key innovation lies in the integration of local spatial, global temporal, and generative modeling within a single unified framework, enabling continuous, data-driven prediction of streamline orientation without reliance on discrete sampling or FOD approximations. Furthermore, we design a new network that integrates both local and global contextual information by combining a convolutional neural network (CNN) to extract fine-grained local diffusion features, and an RNN to capture long-range temporal dependencies along the streamline trajectory. These components are unified within a conditional diffusion model that treats fiber tracking as a generative denoising process. The conditional distribution uses spatial embeddings to encode local context and temporal embeddings to incorporate the global trajectory history. This formulation allows predictions to be conditioned on both local dMRI signals and the global context of the streamline and enables accurate and efficient orientation estimation by leveraging anatomical priors and trajectory history throughout the denoising process. We evaluate the proposed approach on both synthetic and in vivo datasets, demonstrating superior performance in resolving fiber crossings, navigating anatomical bottlenecks, and mitigating the effects of noise, highlighting its potential to advance the accuracy and robustness of tractography.

\section{Methods}
\label{sec:methods}
\begin{figure*}[!t]
    \centering
    \includegraphics[width=1\linewidth]{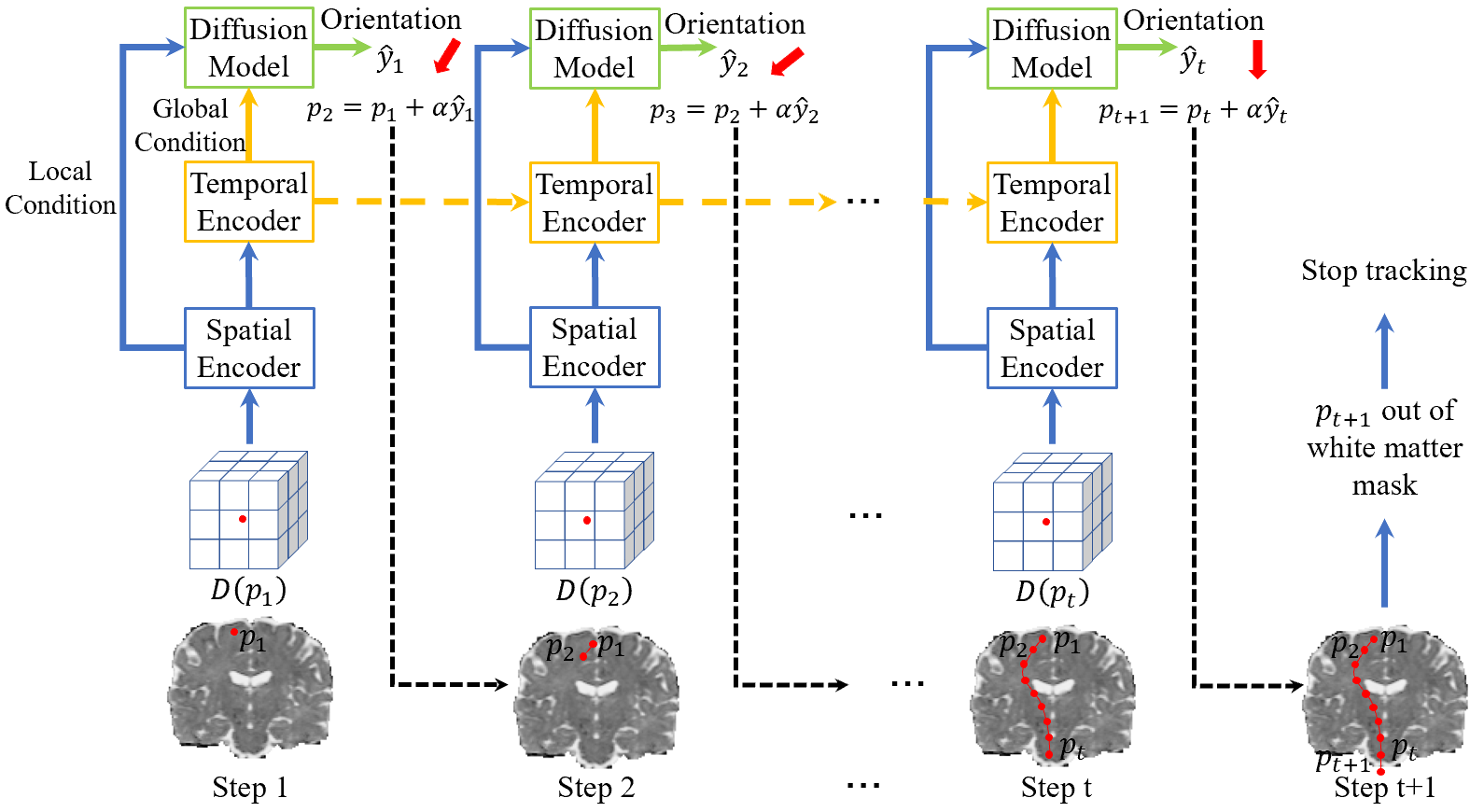}
    \caption{Overview of DDTracking. From the computed SH features at the current position, DDTracking learns the next propagation orientation for fiber tracking. The SH features are passed through the proposed global-local spatiotemporal network, which includes a spatial encoder and a temporal encoder, conditioned on a diffusion model to predict the propagation orientation. This process starts from an initial seed point and repeats iteratively until a stopping criterion is met.}
    \label{fig:methodoverview}
\end{figure*}
The goal of DDTracking is to learn the streamline propagation orientation at each streamline point from input dMRI data and then use the learned orientation to guide fiber tracking. Figure \ref{fig:methodoverview} gives the method overview. Model training data consists of diffusion-weighted imaging (DWI) scans and their corresponding tractograms. For each point along the streamline, spherical harmonic (SH) coefficients are extracted from its local and neighboring voxel DWI signals (Section \ref{subsec:NetworkInput}). As a result, each streamline is represented as an SH feature set, which serves as input to our network. The training process is to iteratively learn the next propagation orientation from the current point position along the streamline. To do so, we design a network architecture that includes: 1) a spatial encoder that captures the local contextual information at each streamline point, 2) a temporal encoder that captures the long-range temporal dependencies of the points along the streamline, and 3) a propagation predictor that integrates the learned spatiotemporal features to predict the next propagation orientation (Section \ref{subsec:NetworkArchitecture}). During inference, the model processes SH coefficients computed from a new DWI scan. Starting from a seed point, the model sequentially estimates a propagation orientation to track fiber pathways, continuing until specified stopping criteria are reached (Section \ref{subsec:FiberTracking}).

\subsection{Network Input}
\label{subsec:NetworkInput}
Our network is trained using two complementary data modalities: input DWI data and the corresponding tractogram as ground truth. Because these two types of data have distinct data structures compared to conventional 2D/3D medical images, a major challenge in our task lies in accurately and efficiently representing tractography streamlines using the underlying DWI signals to enable effective deep neural network processing. In brief, a DWI scan is a 4D image where each 3D volume is acquired under a specific diffusion weighting (b-value) and direction (b-vector) \cite{o2011introduction}. A tractogram consists of a set of streamlines, where each streamline is represented as a sequence of $n$ equally spaced 3D points $S=\begin{Bmatrix}p_1,p_2,\dots,p_n\end{Bmatrix}$. Instead of using the raw 3D coordinates of streamline points, which can primarily capture the streamline geometric properties, we represent each streamline using the local DWI signals to achieve invariance to spatial transformations as in prior work \cite{poulin2017learn,wegmayr2018data}. Specifically, we utilize the SH coefficients \cite{descoteaux2007regularized} derived from the DWI signals for a compact, acquisition-agnostic encoding of diffusion profiles, offering robustness to noise and consistency across dMRI protocols \cite{mirzaalian2016inter,zhu2024diffusion}. For an input DWI scan, each DW image is normalized using the b0 image, followed by projecting the normalized DW images onto an SH basis of degree $m$ \cite{descoteaux2007regularized}, yielding a volume of SH coefficients denoted as $D$, where $m$ determines the number of coefficients. Then, similar to the approach for constructing local signal context along the streamline in \cite{theberge2021track,wegmayr2018data}, we extract a $3\times3\times3$ voxel grid centered at each point  $p$, sampling the SH coefficients from the point itself and its 26 immediate neighbors. This results in a 4D SH feature $D(p)\in \mathbf{R}^{3\times3\times3\times m}$ for each streamline point $p$.

Following the standard streamline propagation process, given a fixed step size $\alpha$ that defines the distance traversed at each step of propagation, from the current point $p_{t\in[1,n]}$, the next point along the streamline is computed as $p_{t+1}=p_t+\alpha y_t$, where $y_t\in\mathbf{R}^3$ denotes the local propagation orientation vector. Then, given the above feature representation of the already tracked points along the streamline, our learning task is to predict $\hat{y}_t$ from the feature set $\begin{Bmatrix}D(p_1),D(p_2),\dots,D(p_t)\end{Bmatrix}$. 
\subsection{Network Architecture}
\label{subsec:NetworkArchitecture}
\begin{figure}
    \centering
    \includegraphics[width=1\linewidth]{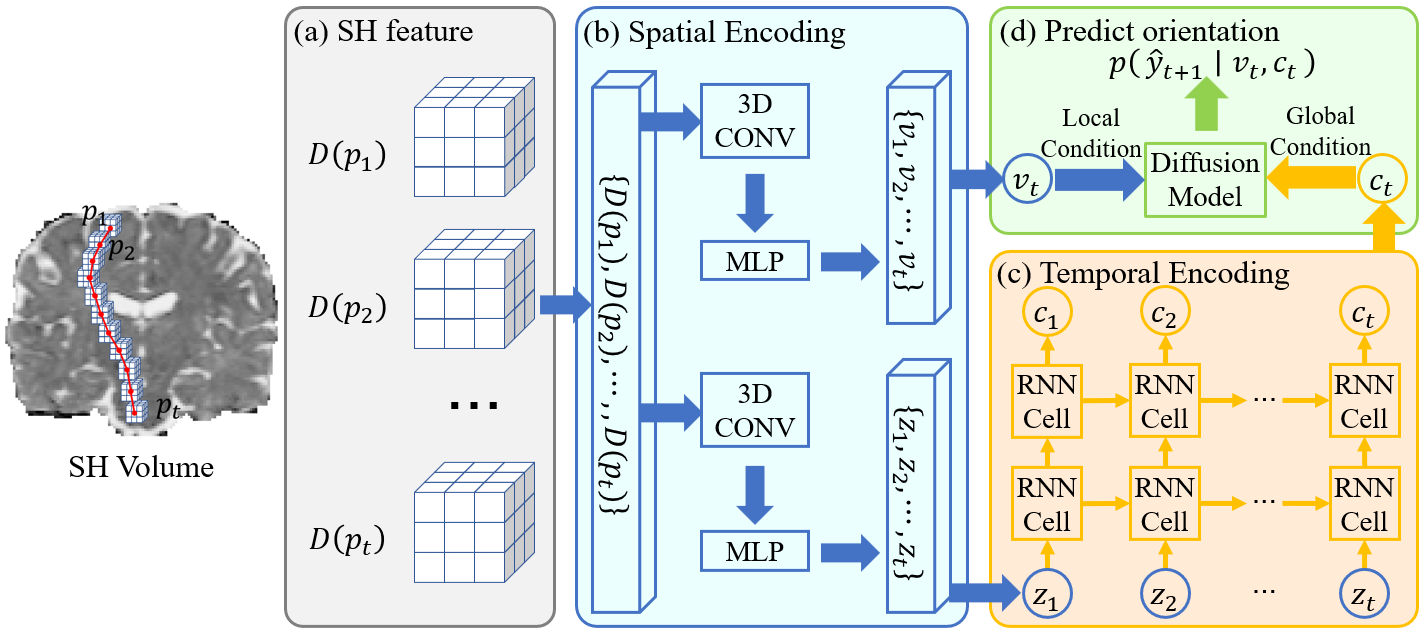}
    \caption{Schematic of the proposed generation framework. The input consists of the SH feature set extracted along a candidate streamline (a). The spatial encoding module (b) extracts two complementary feature representations using dual 3D convolutional (3D CONV) branches followed by MLPs to produce two complementary representations: one capturing local spatial context and the other encoding orientation-aware features. The latter is further processed by the temporal encoding module (c), a stacked RNN that models long-range dependencies along the streamline. These global and local embeddings serve as conditional inputs to a diffusion model (d), which predicts the next orientation step. }
    \label{fig:networkarch}
\end{figure}

Our network architecture to learn streamline propagation orientation for fiber tracking is shown in Figure \ref{fig:networkarch}. It comprises three core components: a CNN for spatial encoding, an RNN for temporal encoding, and a conditioned diffusion model for orientation prediction. The design introduces three key innovations: (1) a dual-branch spatial-temporal encoding scheme that decouples global continuity from localized details, (2) a diffusion-model-based generative framework designed for streamline propagation orientation prediction, and (3) a conditional distribution predictor driven jointly by local spatial features and global temporal context to guide streamline generation.

\subsubsection{Spatial and Temporal Encoding of Diffusion Signals Along Streamlines}
\label{subsubsec:Encoding}
In tractography, both the local neighborhood of a point and the global trajectory of the streamline are critical in accurately modeling white matter pathways. While previous deep learning methods typically focus on one of these aspects in isolation \cite{neher2017fiber,benou2019deeptract,poulin2017learn,theberge2021track,wegmayr2018data}, these two aspects are not exclusive and could benefit from each other. To address this, we adopt a dual-pathway encoding strategy that jointly captures local spatial details and global temporal dynamics.

\textbf{Spatial Encoding:} For each local SH feature $D(p_t)\in \mathbf{R}^{3\times3\times3\times m}$, we apply two separate 3D convolutional projection layers to extract spatial embeddings: $z_t\in \mathbf{R}^{1\times192}$ and $v_t\in \mathbf{R}^{1\times192}$. The embedding $z_t$ flows into the downstream temporal encoder to model the streamline trajectory, preserving the spatial cues' order and orientation continuity along the streamline. In parallel, the embedding $v_t$ serves as the local conditioning input to the diffusion model, capturing localized spatial features at each point. This architectural separation ensures that, during backpropagation, each embedding is independently optimized for its specific role, thereby addressing the complementary demands of streamline tracking. 

\textbf{Temporal Encoding:} The derived spatial embeddings of the points along a streamline, denoted $Z=\begin{Bmatrix}z_1,z_2,\dots,z_t\end{Bmatrix}$, are fed into a two-stacked RNN layer. Specifically, we employ GRUs to compute temporal embeddings. This recurrent structure enables modeling of variable-length streamlines without requiring padding or masking, thus avoiding associated artifacts. At each step, the GRU updates its memory based on the current spatial embedding and accumulated context, resulting in a context-aware temporal embedding $c_t\in \mathbf{R}^{1\times512}$. The representation becomes increasingly informative as it integrates trajectory history and inter-step dependencies of the streamline. Then, this enriched representation serves as a global condition in the diffusion model, enhancing the model’s contextual awareness and enabling anatomically coherent streamline generation.

\subsubsection{Diffusion Model for Prediction of Streamline Propagation Orientation}
\label{subsubsec:DiffusionModel}
\begin{figure}
    \centering
    \includegraphics[width=1\linewidth]{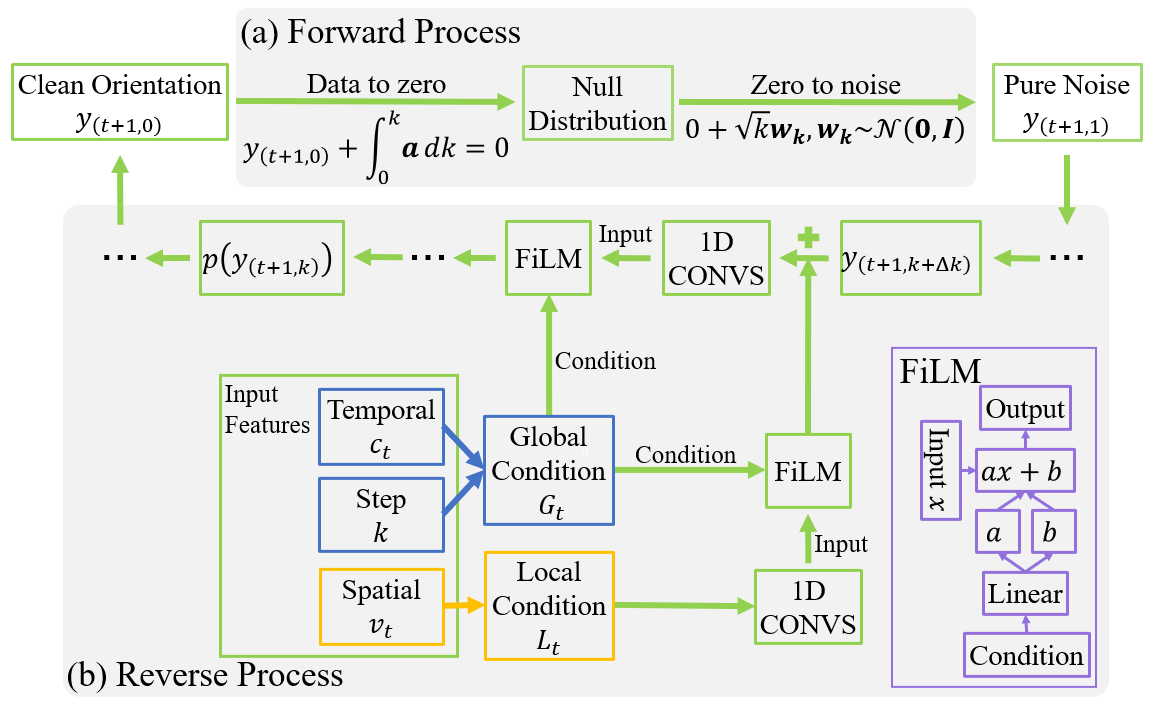}
    \caption{Schematic of the diffusion model for streamline orientation prediction. The upper panel depicts the forward process, where the original (clean) streamline orientation is attenuated to zero while independent Gaussian noise is accumulated, resulting in a smooth transition to a pure noise distribution. The lower panel illustrates the reverse process, which progressively recovers the clean orientation from noise using a 1D CNN with FiLM-modulated residual blocks, guided by global temporal and local spatial conditions. }
    \label{fig:diffusionmodel}
\end{figure}

Next, we design a diffusion model-based module that incorporates the learned spatial and temporal embeddings to predict streamline propagation orientation. Unlike conventional image-based diffusion models that operate with 2D convolutions, our model uses a 1D CNN-based diffusion model \cite{chi2023diffusion} specifically tailored for sequential data to handle streamline sequences. To extend the original single-condition network for global-local spatio-temporal modeling, we adopt a dual-conditioning network that fuses global temporal and local spatial embeddings within a symmetric encoder-decoder architecture based on 1D convolutions, as illustrated in Figure \ref{fig:diffusionmodel}. Specifically, at each diffusion step $k$, the conditional distribution $p(y_{t+1}\mid G_t, L_t)$ is defined using both global and local contexts. The global context $G_t$ combines the temporal embedding $c_t$ with a sinusoidal positional embedding of $k$, while the local context $L_t$ is provided by the spatial embedding $v_t$. These context vectors are fused via Feature-wise Linear Modulation (FiLM) \cite{perez2018film}, which allows for the adaptive modulation of intermediate network activations. 

\textbf{Forward Diffusion Process (Figure \ref{fig:diffusionmodel}(a)):} Unlike standard diffusion models that map data directly to noise, we adopt a decoupled diffusion process comprising two simultaneous sub-processes \cite{lv2024diffmot}: a data-to-zero signal attenuation and a zero-to-noise noise injection. Given a continuous diffusion step $k\in[0,1]$, we simultaneously attenuate the clean orientation data $y_{(t,0)}$ to zero and inject Gaussian noise into the signal. The data-to-zero component employs a constant analytic attenuation function $\boldsymbol{h}=-y_{(t,0)}$, resulting in a smooth linear decay of the original signal over time. Simultaneously, the zero-to-noise component gradually adds standard Gaussian noise $\epsilon$  to the zeroed signal. With this choice, the forward sampling simplifies to: $y_{(t,k)}=y_{(t,0)}+k\boldsymbol{h}+\sqrt{k}\epsilon,\epsilon\sim\mathcal{N}(\boldsymbol{0},\boldsymbol{I})$.

\textbf{Reverse Diffusion Process (Figure \ref{fig:diffusionmodel}(b)):} The reverse process is formulated as a continuous-time Markov chain with an infinitesimal step size $\Delta t\to0^+$ to recover the clean orientation from noisy samples. In our method, we propose to leverage the learned local and global embeddings to guide the reverse diffusion process by using a parameterized model $p_\theta\left(y_{(t,k-\Delta k)}|y_{(t,k)}\right)$ to approximate the posterior distribution  $q\left(y_{(t,k-\Delta k)}|y_{(t,k)},y_{(t,0)}\right)\sim\mathcal{N}(y_{(t,k-\Delta k)};\mu,\Sigma)$ under specific spatiotemporal conditions by applying Bayes' rule. Specifically, the model is conditioned on both the global context $G_t$  and the local context $L_t$, enabling it to predict the attenuation $\boldsymbol{h}_{\theta}(y_{(t,k)},G_t,L_t)$ and noise $\epsilon_{\theta}(y_{(t,k)},G_t,L_t)$ at each diffusion step $k$. The variance is constant and is given as: $\Sigma=\frac{\Delta k(k-\Delta k)}{k}\mathbf{I}$. Following the formulation in DiffMot \cite{lv2024diffmot}, we express noise $\epsilon_{\theta}$ and the mean $\mu_{\theta}$:
\begin{equation}\epsilon_\theta(y_{t,k},G_t,L_t)=\frac{1}{\sqrt{k}}\left[y_{(t,k)}-(k-1)\boldsymbol{h}_\theta(y_{t,k},G_t,L_t)\right],\label{eq2}\end{equation}
\begin{equation}
     \mu_\theta=\frac{k-\Delta k}{k}y_{(t,k)}-\frac{\Delta k}{k}\boldsymbol{h}_\theta(y_{t,k},G_t,L_t).\label{eq3}
\end{equation}
\subsubsection{Training Loss}
\label{subsubsec:loss}
The objective minimizes the expected reconstruction error of the orientation vector over the data and noise components, with the following loss function:
\begin{equation}\min_\theta\mathbb{E}_{q\left(y_{(t,0)}\right)}\mathbb{E}_{q(\epsilon)}[\lambda_1\|\boldsymbol{h_\theta}(y_{t,k},G_t,L_t)_\theta-\boldsymbol{h}\|^2\notag+\lambda_2\|\epsilon_\theta(y_{t,k},G_t,L_t)-\epsilon\|^2].\label{eq3}\end{equation}
To balance their contributions, we introduce dynamic weighting terms: $\lambda_{1}=\frac{k^{2}-k+1}{k}$ and $\lambda_2=\frac{k^2-k+1}{(1-k)^2}$, and apply Smooth L1 loss to ensure training loss robustness.

\begin{table*}[!t]
\caption{{dMRI acquisition and demographics of the datasets studied.}}
\label{tab:DatasetDem}
\setlength{\tabcolsep}{3pt}
\centering
\begin{tabular}{|m{20mm}|m{82mm}|m{70mm}|}
\hline
\makecell[c]{Dataset} & 
\makecell[c]{dMRI Scanner and Acquisition Parameters} & 
\makecell[c]{Demographics and Data Preprocessing}\\
\hline
\makecell[c]{HCP-YA} & 
3T Connectome Siemens Skyra, 90 DWIs @ b=1000s/mm$^2$+18 b=0, TE/TR=89/5520 ms, 1.25mm$\times$1.25mm$\times$1.25mm resolution.&High-quality young adult data; 105 subjects; F/M=60/45; age: 29.3$\pm$3.6.\\ 
\hline
\makecell[c]{ISMRM} &
Simulated, 32 DWIs @ b=1000s/mm$^2$+2 b=0, 2mm$\times$2mm$\times$2mm resolution. &
Synthetic dataset with 25 ground-truth tracts; preprocessed via Tractoflow \cite{theaud2020tractoflow}.\\
\hline
\makecell[c]{TractoInferno} &
Site 1: 3T Philips Achieva, 21 DWIs @ b=1000s/mm$^2$+1 b=0, TE/TR=81/8500ms, 2mm$\times$2mm$\times$2mm resolution;\newline
Site 2: 3T Siemens Prisma, 32 DWIs @ b=1000s/mm$^2$+3 b=0, TE/TR=75/3540ms, 1.75mm$\times$1.75mm$\times$1.75mm resolution;\newline
Site 3: 3T Siemens Prisma, 64 DWIs @ b=1000s/mm$^2$+4 b=0, TE/TR=70/1800ms, 2mm$\times$2mm$\times$2mm resolution;\newline
Site 4: 3T Siemens Trio, 64 DWIs @ b=1000s/mm$^2$+1 b=0, TE/TR=93/9000ms, 2mm$\times$2mm$\times$2mm resolution; \newline
Site 5: 3T GE Discovery MR750, 45 DWIs @ b=1000s/mm$^2$+5 b=0, TE/TR=81/7000ms, 2.3mm$\times$2.3mm$\times$2.3mm resolution; \newline
Site 6: 3T Siemens Magnetom TIM Trio, 30 DWIs @ b=700s/mm$^2$+1 b=0, TE/TR=84/9200ms, 2mm$\times$2mm$\times$2mm resolution. &
Multi-site data from six centers; 284 subjects; preprocessed via Tractoflow \cite{theaud2020tractoflow}.\\
\hline
\makecell[c]{PPMI} &
3T Siemens Trio, 64 DWIs @ b=1000 s/mm$^2$+1 b=0, TE/TR=88/7600 ms, 2mm$\times$2mm$\times$2mm resolution. &
Parkinson’s disease and age-matched HC; 20 subjects; F/M=7/13; age: 63.2$\pm$5.8; PD/HC=11/9; preprocessed via pnlpipe \cite{cetin2024harmonized}.\\
\hline
\makecell[c]{ABIDE} &
3T Siemens Allegra, 64 DWIs @ b=1000 s/mm$^2$+1 b=0, TE/TR=78/5200 ms, 3mm$\times$3mm$\times$3mm resolution. &
ASD and HC; 20 subjects; F/M=2:18; age: 12.1$\pm$7.8; ASD/HC=13/7; preprocessed using pnlpipe \cite{cetin2024harmonized}.\\
\hline
\makecell[c]{CNP} &
3T Siemens TrioTim, 64 DWIs @ b=1000 s/mm$^2$+1 b=0, TE/TR=93/9000 ms, 2mm$\times$2 mm$\times$2mm resolution. &
ADHD, BD, SZ, and HC; 20 subjects; F/M=7:13; age: 32.3$\pm$8.9; ADHD/BD/SZ/HC=4/6/4/6; preprocessed via pnlpipe \cite{cetin2024harmonized}.\\
\hline
\makecell[c]{BrainTumor} &
3T Siemens Prisma, 64 DWIs @ b=1000 s/mm$^2$+1 b=0, TE/TR=79/22000 ms, 1.7mm$\times$1.7mm$\times$4.8mm resolution. &
Private brain tumor dataset; 20 subjects; F/M=7/13; age: 51$\pm$13.8; all patients; preprocessed via pnlpipe \cite{cetin2024harmonized}.\\
\hline
\makecell[c]{Stroke} &
3T GE Discovery MR750, 30 DWIs @ b=1000 s/mm$^2$+1 b=0, TE/TR=80/8750 ms, 2mm$\times$2mm$\times$2mm resolution. &
Private stroke-related cohort; 20 subjects; F/M=8/12; age: 59.6$\pm$13.9; all patients; preprocessed via pnlpipe \cite{cetin2024harmonized}.\\
\hline
\multicolumn{3}{p{175mm}}{Abbreviations: HCP-YA - Human Connectome Project Young Adult; PPMI - Parkinson’s Progression Markers Initiative; PD - Parkinson’s disease; HC - Healthy Control; ABIDE - Autism Brain Imaging Data Exchange; ASD - Autism Spectrum Disorder; CNP - Consortium for Neuropsychiatric Phenomics; ADHD - Attention-Deficit/Hyperactivity Disorder; BD - Bipolar Disorder; SZ - Schizophrenia.} \\
\end{tabular}
\end{table*}

\subsection{Fiber Tracking via Predicted Propagation Orientation}
\label{subsec:FiberTracking}
The implementation of the streamline generation process follows a standard deterministic fiber tracking pipeline \cite{poulin2017learn}. Upon acquiring a new DWI dataset, we first compute the SH coefficients at each voxel. Starting from a seed point, the corresponding SH feature is fed into the trained network to predict the propagation orientation vector $\hat{y}_t$. With the step size $\alpha$, the following point is computed and appended to the streamline trajectory. Then the SH feature of this point is fed into the network for the subsequent prediction. This process repeats until a stopping criterion, defined by either a curvature threshold or a predefined termination mask, is satisfied.  

\subsection{Implementation and Parameter Settings}
\label{subsec:Implementation}
Model training is conducted on a Linux workstation equipped with 16 DIMMs (16 $\times$ 32 GB RAM) and 6 NVIDIA GeForce RTX 4090 GPUs (24 GB memory) using PyTorch v2.2.2. Optimization employed the AdamW algorithm with an initial learning rate of 1e-4. A learning rate decay schedule is adopted: if the validation loss does not improve for 50 consecutive epochs, the learning rate is reduced by a factor of 10, with a floor of 1e-7. Early stopping is applied, with a patience of 120 epochs, to mitigate overfitting. The diffusion signal is represented using an SH basis expansion up to a maximum order of $l_{max}$=6, resulting in $m$=28 coefficients per voxel. For tractography, streamline propagation is initialized with five seeds per voxel within the white matter mask, with a step size of one voxel, and stopped when reaching outside of the mask or a curvature-based angular threshold of 45 degrees.

\section{Experiments and Results}
\label{sec:ExperimentsAndReesults}
\subsection{Datasets and Data Preprocessing}
\label{subsec:Datasets}
We evaluate DDTracking on a diverse set of independently acquired dMRI datasets, including: 1) the Human Connectome Project Young Adult (HCP-YA) \cite{van2013wu}, 2) the ISMRM 2015 Tractography Challenge dataset \cite{maier2017challenge}, 3) TractoInferno \cite{poulin2022tractoinferno}, 4) the Parkinson’s Progression Markers Initiative (PPMI) \cite{marek2011parkinson}, 5) the Autism Brain Imaging Data Exchange (ABIDE) \cite{di2017enhancing},  6) the Consortium for Neuropsychiatric Phenomics (CNP) \cite{poldrack2016phenome}, 7) a brain tumor patient dataset (BrainTumor) acquired at the First Affiliated Hospital of Sun Yat-sen University, China, and 8) a stroke patient dataset (Stroke) acquired at Nuclear Industry 215 Hospital of Shaanxi Province, China. The MRI datasets comprise synthetic and real-world clinical imaging data, spanning different age groups (teenagers to elderly adults), multiple health conditions (brain disorder and neurosurgical patients), and varying acquisition protocols and scanners (see Table \ref{tab:DatasetDem}).

Each dataset undergoes standardized preprocessing pipelines to ensure consistency across acquisition protocols. Specifically, for the HCP-YA, ISMRM, and TractoInferno, we use dMRI data that were processed and provided by the dataset owners, as described in \cite{van2013wu}, \cite{maier2017challenge}, and \cite{poulin2022tractoinferno}, respectively. For PPMI, ABIDE, CNP, BrainTumor, and Stroke, we process the dMRI data using a well-established pipeline (\underline{https://github.com/pnlbwh/pnlpipe}) \cite{cetin2024harmonized}, including brain extraction, eddy current correction, EPI distortion correction, rigidly registering to MNI space, and spatial resolution sampling to 1.25mm$\times$1.25mm$\times$1.25mm (consistent with HCP-YA data that has the highest spatial resolution under study). Then, SH representations are computed for all datasets using DIPY \cite{garyfallidis2014dipy} with the $Descoteaux07$ basis \cite{descoteaux2007regularized}. We train our model on the HCP-YA dataset, utilizing high-quality white matter reference tracts derived using iFOD2 tractography \cite{tournier2010improved} provided by the TractSeg project \cite{wasserthal2018tractseg}. In line with prior studies \cite{neher2017fiber,poulin2017learn,yang2025deep}, ten subjects are used to train the model. This model is subsequently used in the ablation study (Section \ref{subsubsec:AblationStudy}), ISMRM benchmark (Section \ref{subsubsec:Benchmark}), and generalization experiments (Section \ref{subsubsec:Generalization}).

\subsection{Experimental Design}
\label{subsec:ExperimentalDesign}
We perform three experiments to evaluate the proposed DDTracking method. First, we perform ablation studies to assess the effectiveness of the key components of the proposed network (Section \ref{subsubsec:AblationStudy}). Second, we compare our method with several state-of-the-art methods on two widely used benchmark datasets (Section \ref{subsubsec:Benchmark}). Third, we evaluate our method’s generalization ability on dMRI data from different populations and acquisition protocols (Section \ref{subsubsec:Generalization}).

\subsubsection{Ablation study}
\label{subsubsec:AblationStudy}
The ablation study assesses the contributions of each component within our network. We evaluate four model variants: (1) $m_{\text{temporal}}$, a model comprising only two stacked GRU layers without the spatial encoder, (2) $m_{\text{spatial+temporal}}$, a model with the CNN module and GRU layers but excluding the generative component, (3) $m_{\text{spatial+temporal+generative}}$, a model equiped with the spatial and temporal encoders as well as the generative module but excluding the local conditioning module, and (4) the complete DDTracking model. All models are trained on data from 10 HCP-YA subjects, with the same loss (Section \ref{subsubsec:loss}) and related parameter settings (Section \ref{subsec:Implementation}), and evaluated on 5 unseen HCP-YA testing subjects. To compare the results, we parcellate the predicted tractography according to the ORG atlas \cite{zhang2018anatomically} that contains a fine-scale parcellation of the entire white matter into 800 fiber clusters and an anatomical fiber tract parcellation of 74 major anatomical fiber tracts. Cluster- and tract-level detection rates are computed to assess a model’s success in reconstructing fiber clusters/tracts \cite{zhang2020deep,xue2023superficial}. Furthermore, we compare the spatial overlap between the predicted fiber tracts and those derived using a widely used reference tractography method. We employ the Unscented Kalman Filter (UKF) tractography \cite{farquharson2013white,vos2013multi} that is highly consistent for fiber tracking in dMRI data from independently acquired populations across ages, health conditions, and dMRI acquisitions \cite{zhang2018anatomically,zhang2020deep,li2024diffusion}. The weighted Dice (wDice) coefficient \cite{cousineau2017test,zhang2019test}, a metric designed specifically for measuring tract spatial overlap, between the predicted and reference tracts is computed.
\begin{table}[!t]
\caption{Ablation study results.}
\label{tab:ablationresult}
\setlength{\tabcolsep}{3pt}
\centering
\begin{tabular}{|c|c|c|c|}\hline
\makecell[c]{Method} &
\makecell[c]{Cluster Detection\\Rate $\uparrow$} &
\makecell[c]{Tract Detection\\Rate $\uparrow$} &
\makecell[c]{wDice $\uparrow$} \\
\hline
$m_{\text{temporal}}$ &
93.73\% &
100\% &
0.71$\pm$0.19 \\
\hline
$m_{\text{spatial+temporal}}$ &
94.69\% &
100\% &
0.74$\pm$0.16 \\
\hline
$m_{\text{spatial+temporal+generavtive}}$ &
95.11\% &
100\% &
0.77$\pm$0.17 \\
\hline
$m_{\text{DDTracking}}$ &
\textbf{95.62\%} &
\textbf{100\%} &
\textbf{0.80$\pm$0.16} \\
\hline
\end{tabular}
\end{table}

The ablation results in Table \ref{tab:ablationresult} show a consistent improvement in cluster-level detection rate as additional components are integrated into the model. In contrast, the tract-level detection rate stays the same across all models. This shows that although all models successfully identify the major fiber tracts, differences in fine-scale cluster detection highlight each module's role in enhancing sensitive fiber tracking. Furthermore, the spatial overlap (wDice) between tracts identified by our method and the reference method consistently improves as additional components are integrated. Taken together, these results show the benefit of incorporating complementary local and global modeling to enhance anatomical fidelity and consistency in fiber tracking.

\subsubsection{Evaluation on benchmark datasets}
\label{subsubsec:Benchmark}
We evaluate our method on two standard benchmark datasets: ISMRM \cite{maier2017challenge} and TractoInferno \cite{poulin2022tractoinferno}, which have been widely used to assess the performance of fiber tracking methods \cite{neher2017fiber,poulin2017learn,wegmayr2021entrack,theberge2021track,yang2025deep}. The ISMRM dataset includes a synthetic dMRI dataset associated with ground-truth fiber tracts. Evaluation is performed using the Tractometer tool \cite{maier2017challenge,cote2013tractometer}, which provides metrics at both streamline and volume levels. In brief, Valid Connections (VC), Invalid Connections (IC), and No Connections (NC) are streamline-based metrics, where VC measures true positive streamlines correctly linking valid anatomical regions, IC measures false positive streamlines connecting invalid regions, and NC counts prematurely terminated streamlines that fail to link any regions. Overlap (OL), Overreach (OR), and F1 score are volume-oriented metrics, where OL measures shared voxels between reconstructed and ground truth tracts, OR measures excess voxels in reconstructed tracts but not present in ground truth, and F1 measures the overall spatial overlap between the reconstruction and the ground truth tracts. The TractoInferno dataset includes a total of 284 dMRI volumes acquired from six acquisition sites with varying protocols (see Table \ref{tab:DatasetDem}). These DWI volumes are pre-splitted by the dataset organizers, with 198 for model training, 58 for validation, and 28 for testing. Performance on the testing data is assessed using the associated TractoEval tool \cite{poulin2022tractoinferno,garyfallidis2018recognition}, which provides performance metrics including OL, OR, and Dice score (similar to the F1 score to measure the spatial overlap between tracts).

We compare our approach against the following SOTA methods, including: 1) Neher et al. \cite{neher2017fiber}, a pioneering method using random forest for fiber tracking, which is the first machine learning based tractography method; 2) Learn to Track \cite{poulin2017learn}, which uses RNNs to capture high-order dependencies along streamlines to improve tract coverage and reduce false positives; 3) Entrack \cite{wegmayr2021entrack}, which adopts probabilistic modeling to account for the uncertainty during tracking; 4) Track-to-Learn \cite{theberge2021track}, which formulates tractography as a reinforcement learning problem to learn tracking policies that generalize beyond supervised traning; and 5) Yang et al. \cite{yang2025deep} that is a transformer-based model incorporating spatial and anatomical priors to improve tracking in anatomically challenging regions.
\begin{table}
\caption{Comparison on the ISMRM benchmark dataset.}
\centering
\label{tab:ISMRMresult}
\begin{tabular}{|c|c|c|c|c|c|c|}\hline
\multirow{2}{*}{Method} &
\multicolumn{3}{c|}{Connections (\%)} &
\multicolumn{3}{c|}{Average tract (\%)} \\
\cline{2-7}
 &
VC $\uparrow$ &
IC $\downarrow$ &
NC $\downarrow$ &
OL $\uparrow$ &
OR $\downarrow$ &
F1 $\uparrow$ \\
\hline
Neher et al. &
38 &
N/A &
N/A &
46 &
\textbf{34} &
N/A \\
\hline
Learn To Track &
49.8 &
43.1 &
7.1 &
65.6 &
N/A &
N/A \\
\hline
Entrack &
52 &
N/A &
N/A &
58 &
\underline{39} &
54 \\
\hline
Track-to-Learn &
\underline{68.5} &
30.7 &
\underline{0.8} &
\underline{65.8} &
N/A &
N/A \\
\hline
Yang et al. &
66.2 &
\underline{28.3} &
5.5 &
63.8 &
42.3 &
\textbf{56.6} \\
\hline
DDTracking &
\textbf{74.1} &
\textbf{25.9} &
\textbf{0.001} &
\textbf{70.8} &
44.3 &
\underline{56.2} \\
\hline
\multicolumn{7}{p{100mm}}{\textbf{Bold} and \underline{underline} indicate the 1st and 2nd ranking results, respectively.}
\end{tabular}
\end{table}

\begin{table}
\caption{Comparison on the TractoInferno benchmark dataset.}
\centering
\label{tab:TractoInfernoresult}
\begin{tabular}{|c|c|c|c|}
\hline
\makecell[c]{Method} &
\makecell[c]{Dice $\uparrow$} &
\makecell[c]{Overlap $\uparrow$} &
\makecell[c]{Overreach $\downarrow$} \\
\hline
\makecell[c]{Learn To Track} &
\makecell[c]{58.0} &
\makecell[c]{53.3} &
\makecell[c]{25.8} \\
\hline
\makecell[c]{Track-To-Learn} &
\makecell[c]{48.4} &
\makecell[c]{35.1} &
\makecell[c]{\textbf{5.4}} \\
\hline
\makecell[c]{Yang et al.} &
\makecell[c]{\textbf{64.5}} &
\makecell[c]{\underline{59.0}} &
\makecell[c]{\underline{21.7}} \\
\hline
\makecell[c]{DDTracking} &
\makecell[c]{\textbf{64.5}} &
\makecell[c]{\textbf{64.6}} &
\makecell[c]{36.0} \\
\hline
\multicolumn{4}{p{100mm}}{\textbf{Bold} and \underline{underline} indicate the 1st and 2nd ranking results, respectively.}
\end{tabular}
\end{table}

Table \ref{tab:ISMRMresult} gives the evaluation results on the ISMRM dataset. Because the ISMRM dataset provides only one DWI volume, to avoid potential data leakage and overfitting, all models are trained on the HCP-YA dataset and evaluated on ISMRM. Scores for Neher et al., Learn to Track, Entrack, and Track-to-Learn are extracted from the Track-to-Learn paper, while results for Yang et al. are taken from their original publication. Our method achieves the best performance in VC, IC, NC, and OL, and the second-best performance in the F1 metric, while achieving moderate performance in OR (44.3\%). 

Table \ref{tab:TractoInfernoresult} gives the evaluation results on the TractoInferno dataset. All compared methods are trained on the TractoInferno training set, and evaluation metrics are computed based on the testing set. Performance scores for Learn to Track, Track-to-Learn, and the method by Yang et al. are extracted from Yang et al.’s publication. Results of the Neher et al. and Entrack are not available in the TractoInferno dataset. Our method achieves the highest reported Dice score (64.5\%, equal to Yang et al.) and the highest Overlap score (64.6\%), with a moderate increase in Overreach (36.0\%).

\subsubsection{Evaluation on generalization to other datasets}
\label{subsubsec:Generalization}
This experiment assesses the generalization capability of our method in real-world scenarios across a diverse set of dMRI datasets, encompassing diverse clinical populations and acquisition settings, including HCP-YA, PPMI, ABIDE, CNP, BrainTumor and Stroke (see Table \ref{tab:DatasetDem}). In all cases, the model is trained on the HCP-YA dataset and tested on each of the aforementioned testing datasets to predict whole-brain tractography. Unlike the benchmark datasets used above, there are no ground truth tractography results on the datasets used in this experiment. Therefore, the goal here is to evaluate whether DDTracking can generate anatomically comparable results to the existing tractography methods. We compare with three existing methods, including: 1) UKF tractography \cite{malcolm2010filtered,reddy2016joint} that performs whole-brain tractography by fitting a two-tensor model from the diffusion signals using the Unscented Kalman Filter approach, 2) iFOD2 \cite{tournier2010improved} that is a probabilistic method based on second-order integration over fiber orientation distributions, and 3) TractSeg \cite{wasserthal2018tractseg} that is a tract-specific tractography method that leverages deep learning for bundle segmentation and streamline generation. Parameters per method are set to the suggested settings in their software packages. 

For quantitative evaluation, we extract anatomical fiber tracts from the tractography results and compute both the tract detection rate and the spatial overlap (using wDice) between our method and each compared approach. Since only TractSeg is a tract-specific method that directly generates anatomical tracts, we standardize comparisons by parcellating whole-brain tractography from other methods using TractSeg-derived tract segmentation and endpoint masks as region-of-interest (ROI) constraints. Specifically, extracted tracts must lie entirely within the corresponding TractSeg tract mask, with both endpoints confined to their respective endpoint masks.
\begin{table*}[!t]
\caption{Tract detection rate and spatial overlap between the four methods.}
\label{tab:generalizationresult}
\centering
\begin{tabular}{|c|c|c|c|c|c|c|c|}
\hline
\multirow{2}{*}{Dataset} &
\multicolumn{4}{c|}{Tract Detection Rate} &
\multicolumn{3}{c|}{Spatial Overlap (wDice)} \\
\cline{2-8}
 &
UKF &
iFOD2 &
TractSeg &
DDTracking &
\makecell[c]{UKF vs \\DDTracking} &
\makecell[c]{iFOD2 vs \\DDTracking} &
\makecell[c]{TractSeg vs \\DDTracking} \\
\hline
HCP-YA &
98\% &
100\% &
100\% &
100\% &
0.73 $\pm$ 0.19 &
0.82 $\pm$ 0.15 &
0.75 $\pm$ 0.14 \\
\hline
PPMI &
98\% &
100\% &
100\% &
99\% &
0.82 $\pm$ 0.17 &
0.87 $\pm$ 0.14 &
0.83 $\pm$ 0.14 \\
\hline
ABIDE &
97\% &
99\% &
100\% &
99\% &
0.88 $\pm$ 0.15 &
0.92 $\pm$ 0.12 &
0.89 $\pm$ 0.12 \\
\hline
CNP &
92\% &
95\% &
94\% &
94\% &
0.82 $\pm$ 0.18 &
0.87 $\pm$ 0.16 &
0.82 $\pm$ 0.15 \\
\hline
BrainTumor &
93\% &
97\% &
96\% &
95\% &
0.77 $\pm$ 0.23 &
0.87 $\pm$ 0.16 &
0.82 $\pm$ 0.17 \\
\hline
Stroke &
78\% &
81\% &
79\% &
81\% &
0.71 $\pm$ 0.17 &
0.72 $\pm$ 0.20 &
0.72 $\pm$ 0.20 \\
\hline
\end{tabular}
\end{table*}

\begin{figure*}[!t]
    \centering
    \includegraphics[width=1\linewidth]{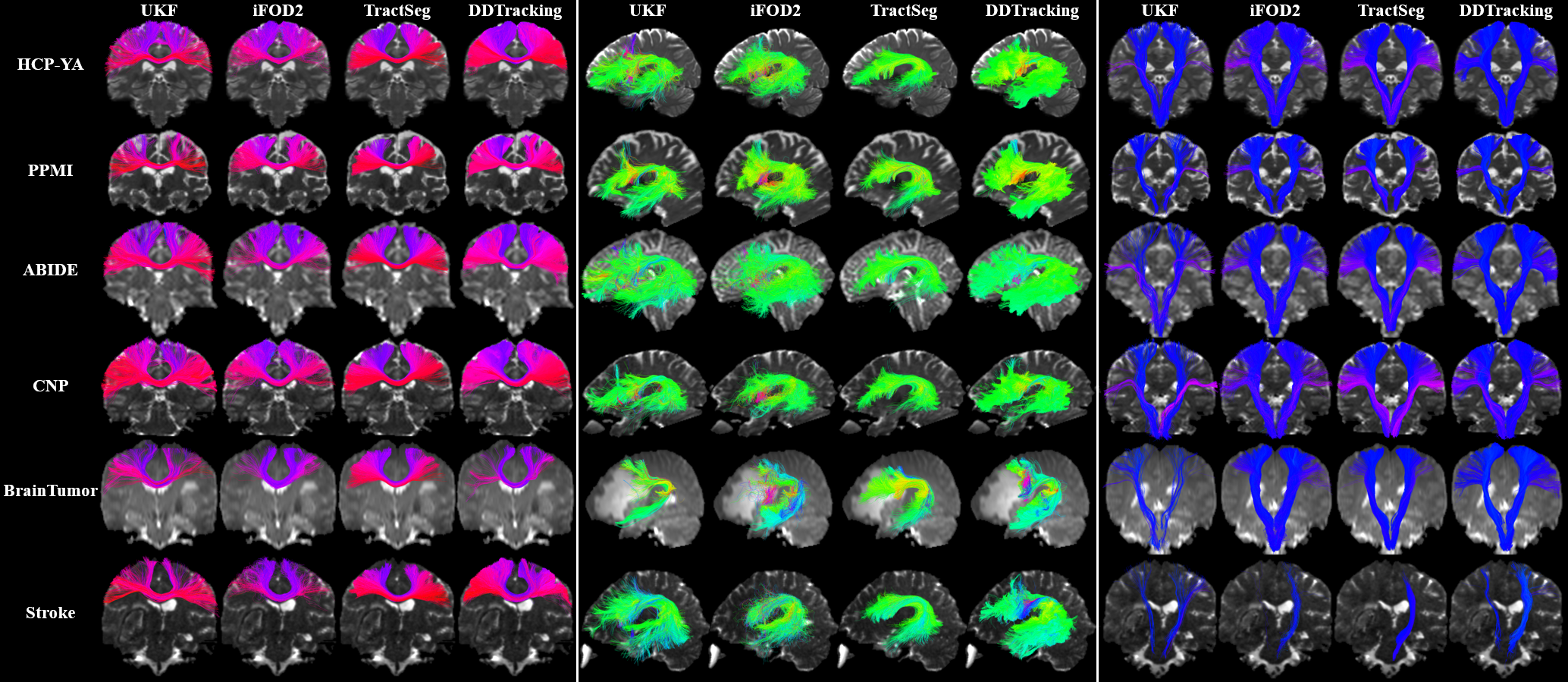}
    \caption{Qualitative comparison of tractography results on multiple datasets in several example tracts (CC4, AF, and CST).}
    \label{fig:generalizationfig}
\end{figure*}

Table \ref{tab:generalizationresult} presents quantitative comparisons across all datasets. Our method achieves comparable tract detection rates, indicating reliable identification of the majority of white matter tracts across diverse populations. In addition, our approach attains high and competitive spatial overlap scores, as measured by the wDice metric, in comparison with TractSeg, iFOD2, and UKF. A wDice score over 0.72 is suggested to be a good tract overlap  \cite{cousineau2017test,zhang2019test}. These results show that the streamlines generated by our method exhibit anatomical consistency and align well with those produced by widely used tractography pipelines. 

\begin{figure*}[!t]
    \centering
    \includegraphics[width=0.5\linewidth]{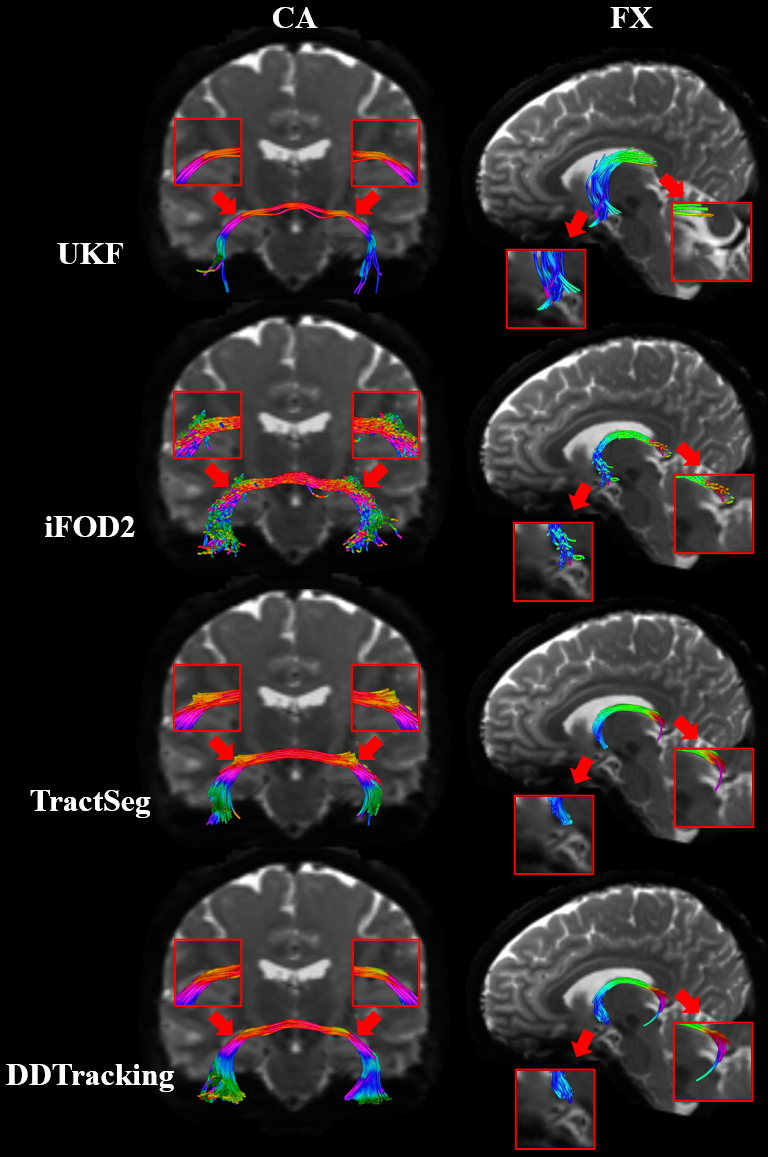}
    \caption{Comparison of hard-to-track tracts (CA and FX) in an example HCP-YA test subject.}
    \label{fig:hardtotrack}
\end{figure*}

Figure \ref{fig:generalizationfig} provides qualitative visualizations of three example fiber tracts in randomly selected subjects from each dataset. The extracted tract shapes in our method closely resemble those produced by the baseline methods. Figure \ref{fig:hardtotrack} gives a further visual comparison of two representative, hard-to-track tracts - the anterior commissure (CA) and fornix (FX). Our method successfully reconstructs the tracts and achieves continuous tracking in challenging regions (as indicated by red arrows), whereas the compared methods extended beyond the expected region (e.g., FX in UKF) or generated broken tracts (e.g., CA in iFod2 and TractSeg). Overall, the visual results demonstrate that while DDTracking achieves generally comparable performance to existing methods, it can improve tracking accuracy in locally challenging regions.

\section{Discussion}
\label{sec:Discussion}
We propose a novel deep generative framework for dMRI tractography, which, to our knowledge, is the first work leveraging diffusion models for fiber tracking. Unlike traditional tractography methods that rely on handcrafted models or discrete orientation sampling \cite{malcolm2010filtered,reddy2016joint,aganj2009odf,tournier2012mrtrix,tournier2010improved}, our approach is a learning-based method to automatically learn complex patterns from dMRI data in a data-driven manner, reducing reliance on manually designed models. Furthermore, we reframe streamline tracking as a generative denoising task, leveraging diffusion models to learn the underlying streamline propagation distributions to capture the inherent variability and complexity of white matter fiber pathways. Our experiments on two widely used benchmark datasets (i.e., ISMRM and TractoInferno) show that DDTracking outperforms several existing state-of-the-art tractography methods. In related work, Entrack \cite{wegmayr2021entrack} proposes a general probabilistic model for spherical regression to learn local streamline orientations, and similarly, Track-to-Learn \cite{theberge2021track} applies reinforcement learning to predict fiber tracking orientations from local neighboring information. In contrast, Learn to Track \cite{poulin2017learn} adopts an RNN-based model to focus mainly on the global dependency of points along the streamline.  More recently, a transformer-based model is proposed to learn global dependency along the streamline from local contextual information \cite{yang2025deep}. In our DDTracking method, we propose a dual-pathway encoding network that jointly models local spatial encoding and global temporal dependencies for improved fiber tracking. 

We demonstrate the complementary roles of local and global modeling for improved tractography. Unlike prior approaches that focus on either local or global context \cite{neher2017fiber,poulin2017learn,theberge2021track}, DDTracking integrates both local neighborhood structure and long-range trajectory information to capture the complementary spatial and temporal features for streamline reconstruction. Consistent with previous work using recurrent or memory-based mechanisms to model the global trajectory context \cite{benou2019deeptract,poulin2017learn}, our result shows that using a recurrent network ($m_{\text{temporal}}$) can obtain good cluster-level and tract-level detection rates (93.7\% and 100\%, respectively) and a reasonably good spatial overlap with the widely used reference tractography method (wDice=0.71). However, relying solely on temporal modeling may be insufficient in capturing local variations critical in regions with complex fiber configurations, a limitation highlighted in prior studies \cite{neher2017fiber,wegmayr2018data}. Our results show that incorporating local spatial encoding ($m_{\text{spatial+temporal}}$) largely improves the wDice scores over temporal modeling alone (0.74 vs 0.71). This improvement is further amplified when integrating a local conditioning module ($m_{\text{DDTracking}}$) into the generative process (wDice=0.80), outperforming the compared model without the local conditioning module ($m_{\text{spatial+temporal+generative}}$) (wDice=0.77). Overall, we show that combining these complementary perspectives can achieve anatomically coherent streamline reconstruction that remains locally adaptive yet globally consistent across the entire trajectory.

DDTracking demonstrates a good generalizability across dMRI datasets spanning diverse populations, age groups, imaging protocols, and scanner types. Our trained model benefits from the high-quality HCP-YA data, leveraging its rich spatial and angular resolution. Consistent with prior studies \cite{theberge2021track,theberge2024matters,zhang2020deep,zhang2021deep}, the strategy of training on high-quality dMRI data enables effective transfer to heterogeneous datasets. In addition, we adopt the SH representation of the dMRI signals. This representation decouples the model from specific acquisition parameters (e.g., b-values, gradient directions), enabling uniform processing of heterogeneous data \cite{mirzaalian2016inter,zhu2024diffusion}. Furthermore, rather than depending on predefined physical models, the proposed network learns a flexible, data-derived representation of white matter geometry to model the distribution of feasible trajectories, increasing robustness across diverse anatomical and acquisition variability. Importantly, we show that DDTracking enables reliable fiber tracking in populations with large anatomical variability, including patients with brain tumors and stroke. Fiber tracking is particularly challenging in such patients, where white matter anatomy largely deviates from healthy populations \cite{golby2011interactive,leclercq2011diffusion,yamada2009mr,yu2005diffusion,essayed2017white}. Overall, the good cross-dataset performance on clinically diverse cohorts in multiple brain diseases demonstrates the robust generalization capabilities of DDTracking. 

Potential future directions and limitations of the current work are as follows. First, DDTracking operates solely on dMRI data without incorporating anatomical priors. The integration of such priors, commonly employed in methods like Anatomically-Constrained Tractography (ACT) \cite{smith2012anatomically}, could enhance streamline termination accuracy and improve tract specificity. Second, although we use GRUs to capture sequential dependencies, other advanced architectures, such as Transformers \cite{vaswani2017attention}, Mambas \cite{gu2023mamba}, or other emerging sequence models, may offer greater capacity for long-range context modeling and richer latent representations. Third, our model operates on SH coefficients, which are primarily suited for single-shell data. Although single-shell signals can be extracted from multi-shell acquisitions, fully exploiting the richness of the multi-shell dMRI signals may benefit from incorporating more expressive signal representations. Finally, in our current study, we employ conventional deterministic fiber tracking by predicting a single propagation orientation vector at each position. However, our method can be extended to probabilistic fiber tracking by predicting multiple orientations, enabling more advanced fiber tracking strategies. 
\section{Conclusion}
\label{sec:Conclusion}
In this work, we propose DDTracking, a novel deep generative dMRI tractography framework that integrates local spatial encoding, global temporal modeling, and a conditional diffusion model-based generative process to predict streamline propagation orientations. By integrating convolutional and recurrent modules within a probabilistic generative process, our method effectively captures both local dMRI signals and long-range trajectory dependencies without relying on predefined orientation distributions or discrete sampling. Through evaluations on synthetic, benchmark, and clinical datasets, DDTracking shows good generalization, maintaining anatomical plausibility and robustness across diverse acquisition conditions and anatomical variations. Overall, our results show that DDTracking produces anatomically plausible, robust tractography, offering a scalable, adaptable, and learnable solution for dMRI applications.

\section*{Acknowledgments}

This work is in part supported by the National Key R\&D Program of China (No. 2023YFE0118600) and the National Natural Science Foundation of China (No. 62371107). 

\bibliography{main}

\begin{thebibliography}{10}
\urlstyle{rm}
\expandafter\ifx\csname url\endcsname\relax
  \def\url#1{\texttt{#1}}\fi
\expandafter\ifx\csname urlprefix\endcsname\relax\def\urlprefix{URL }\fi
\expandafter\ifx\csname doiprefix\endcsname\relax\def\doiprefix{DOI: }\fi
\providecommand{\bibinfo}[2]{#2}
\providecommand{\eprint}[2][]{\url{#2}}

\bibitem{basser1994mr}
\bibinfo{author}{Basser, P.}, \bibinfo{author}{Mattiello, J.} \& \bibinfo{author}{LeBihan, D.}
\newblock \bibinfo{journal}{\bibinfo{title}{{MR diffusion tensor spectroscopy and imaging}}}.
\newblock {\emph{\JournalTitle{Biophysical journal}}} \textbf{\bibinfo{volume}{66}}, \bibinfo{pages}{259--267} (\bibinfo{year}{1994}).

\bibitem{basser2000vivo}
\bibinfo{author}{Basser, P.}, \bibinfo{author}{Pajevic, S.}, \bibinfo{author}{othersPierpaoli, C.}, \bibinfo{author}{Duda, J.} \& \bibinfo{author}{Aldroubi, A.}
\newblock \bibinfo{journal}{\bibinfo{title}{{In vivo fiber tractography using DT-MRI data}}}.
\newblock {\emph{\JournalTitle{Magnetic resonance in medicine}}} \textbf{\bibinfo{volume}{44}}, \bibinfo{pages}{625--632} (\bibinfo{year}{2000}).

\bibitem{zhang2022quantitative}
\bibinfo{author}{Zhang, F.} \emph{et~al.}
\newblock \bibinfo{journal}{\bibinfo{title}{{Quantitative mapping of the brain’s structural connectivity using diffusion MRI tractography: A review}}}.
\newblock {\emph{\JournalTitle{Neuroimage}}} \textbf{\bibinfo{volume}{249}}, \bibinfo{pages}{118870} (\bibinfo{year}{2022}).

\bibitem{o2011introduction}
\bibinfo{author}{O’Donnell, L.} \& \bibinfo{author}{Westin, C.-F.}
\newblock \bibinfo{journal}{\bibinfo{title}{{An introduction to diffusion tensor image analysis}}}.
\newblock {\emph{\JournalTitle{Neurosurgery Clinics of North America}}} \textbf{\bibinfo{volume}{22}}, \bibinfo{pages}{185} (\bibinfo{year}{2011}).

\bibitem{malcolm2010filtered}
\bibinfo{author}{Malcolm, J.}, \bibinfo{author}{Shenton, M.} \& \bibinfo{author}{Rathi, Y.}
\newblock \bibinfo{journal}{\bibinfo{title}{{Filtered multitensor tractography}}}.
\newblock {\emph{\JournalTitle{IEEE transactions on medical imaging}}} \textbf{\bibinfo{volume}{29}}, \bibinfo{pages}{1664--1675} (\bibinfo{year}{2010}).

\bibitem{reddy2016joint}
\bibinfo{author}{Reddy, C.} \& \bibinfo{author}{Rathi, Y.}
\newblock \bibinfo{journal}{\bibinfo{title}{{Joint multi-fiber NODDI parameter estimation and tractography using the unscented information filter}}}.
\newblock {\emph{\JournalTitle{Frontiers in neuroscience}}} \textbf{\bibinfo{volume}{10}}, \bibinfo{pages}{166} (\bibinfo{year}{2016}).

\bibitem{aganj2009odf}
\bibinfo{author}{Aganj, I.}, \bibinfo{author}{Lenglet, C.} \& \bibinfo{author}{Sapiro, G.}
\newblock \bibinfo{title}{{ODF reconstruction in q-ball imaging with solid angle consideration}}.
\newblock In \emph{\bibinfo{booktitle}{ISBI}}, \bibinfo{pages}{1398--1401} (\bibinfo{year}{2009}).

\bibitem{tournier2012mrtrix}
\bibinfo{author}{Tournier, J.-D.}, \bibinfo{author}{Calamante, F.} \& \bibinfo{author}{Connelly, A.}
\newblock \bibinfo{journal}{\bibinfo{title}{{MRtrix: diffusion tractography in crossing fiber regions}}}.
\newblock {\emph{\JournalTitle{International journal of imaging systems and technology}}} \textbf{\bibinfo{volume}{22}}, \bibinfo{pages}{53--66} (\bibinfo{year}{2012}).

\bibitem{leemans2025deterministic}
\bibinfo{author}{Leemans, A.}, \bibinfo{author}{Dell’Acqua, F.} \& \bibinfo{author}{Descoteaux, M.}
\newblock \bibinfo{title}{{Deterministic fiber tractography}}.
\newblock In \emph{\bibinfo{booktitle}{Handbook of Diffusion MR Tractography}}, \bibinfo{pages}{241--255} (\bibinfo{publisher}{Elsevier}, \bibinfo{year}{2025}).

\bibitem{girard2025probabilistic}
\bibinfo{author}{Girard, G.} \emph{et~al.}
\newblock \bibinfo{journal}{\bibinfo{title}{{Probabilistic tractography}}}.
\newblock {\emph{\JournalTitle{Handbook of Diffusion MR Tractography}}} \bibinfo{pages}{257--274} (\bibinfo{year}{2025}).

\bibitem{descoteaux2007regularized}
\bibinfo{author}{Descoteaux, M.}, \bibinfo{author}{Angelino, E.}, \bibinfo{author}{Fitzgibbons, S.} \& \bibinfo{author}{Deriche, R.}
\newblock \bibinfo{journal}{\bibinfo{title}{{Regularized, fast, and robust analytical Q-ball imaging}}}.
\newblock {\emph{\JournalTitle{Magnetic Resonance in Medicine: An Official Journal of the International Society for Magnetic Resonance in Medicine}}} \textbf{\bibinfo{volume}{58}}, \bibinfo{pages}{497--510} (\bibinfo{year}{2007}).

\bibitem{tournier2004direct}
\bibinfo{author}{Tournier, J.-D.}, \bibinfo{author}{Calamante, F.}, \bibinfo{author}{Gadian, D.} \& \bibinfo{author}{Connelly, A.}
\newblock \bibinfo{journal}{\bibinfo{title}{{Direct estimation of the fiber orientation density function from diffusion-weighted MRI data using spherical deconvolution}}}.
\newblock {\emph{\JournalTitle{Neuroimage}}} \textbf{\bibinfo{volume}{23}}, \bibinfo{pages}{1176--1185} (\bibinfo{year}{2004}).

\bibitem{jeurissen2014multi}
\bibinfo{author}{Jeurissen, B.}, \bibinfo{author}{Tournier, J.-D.}, \bibinfo{author}{Dhollander, T.}, \bibinfo{author}{Connelly, A.} \& \bibinfo{author}{Sijbers, J.}
\newblock \bibinfo{journal}{\bibinfo{title}{{Multi-tissue constrained spherical deconvolution for improved analysis of multi-shell diffusion MRI data}}}.
\newblock {\emph{\JournalTitle{NeuroImage}}} \textbf{\bibinfo{volume}{103}}, \bibinfo{pages}{411--426} (\bibinfo{year}{2014}).

\bibitem{smith2012anatomically}
\bibinfo{author}{Smith, R.}, \bibinfo{author}{Tournier, J.-D.}, \bibinfo{author}{Calamante, F.} \& \bibinfo{author}{Connelly, A.}
\newblock \bibinfo{journal}{\bibinfo{title}{{Anatomically-constrained tractography: improved diffusion MRI streamlines tractography through effective use of anatomical information}}}.
\newblock {\emph{\JournalTitle{Neuroimage}}} \textbf{\bibinfo{volume}{62}}, \bibinfo{pages}{1924--1938} (\bibinfo{year}{2012}).

\bibitem{daducci2016microstructure}
\bibinfo{author}{Daducci, A.}, \bibinfo{author}{Dal~Pal{\'u}, A.}, \bibinfo{author}{Descoteaux, M.} \& \bibinfo{author}{Thiran, J.-P.}
\newblock \bibinfo{journal}{\bibinfo{title}{{Microstructure informed tractography: pitfalls and open challenges}}}.
\newblock {\emph{\JournalTitle{Frontiers in neuroscience}}} \textbf{\bibinfo{volume}{10}}, \bibinfo{pages}{247} (\bibinfo{year}{2016}).

\bibitem{poulin2019tractography}
\bibinfo{author}{Poulin, P.}, \bibinfo{author}{J{\"o}rgens, D.}, \bibinfo{author}{Jodoin, P.-M.} \& \bibinfo{author}{Descoteaux, M.}
\newblock \bibinfo{journal}{\bibinfo{title}{{Tractography and machine learning: Current state and open challenges}}}.
\newblock {\emph{\JournalTitle{Magnetic resonance imaging}}} \textbf{\bibinfo{volume}{64}}, \bibinfo{pages}{37--48} (\bibinfo{year}{2019}).

\bibitem{karimi2024diffusion}
\bibinfo{author}{Karimi, D.} \& \bibinfo{author}{Warfield, S.}
\newblock \bibinfo{journal}{\bibinfo{title}{{Diffusion MRI with machine learning}}}.
\newblock {\emph{\JournalTitle{Imaging Neuroscience}}} \textbf{\bibinfo{volume}{2}}, \bibinfo{pages}{1--55} (\bibinfo{year}{2024}).

\bibitem{zhang2025think}
\bibinfo{author}{Zhang, F.}, \bibinfo{author}{Th{\'e}berge, A.}, \bibinfo{author}{Jodoin, P.-M.}, \bibinfo{author}{Descoteaux, M.} \& \bibinfo{author}{O’Donnell, L.}
\newblock \bibinfo{journal}{\bibinfo{title}{{Think deep in the tractography game: deep learning for tractography computing and analysis}}}.
\newblock {\emph{\JournalTitle{Brain Structure and Function}}} \textbf{\bibinfo{volume}{230}}, \bibinfo{pages}{100} (\bibinfo{year}{2025}).

\bibitem{neher2017fiber}
\bibinfo{author}{Neher, P.}, \bibinfo{author}{C{\^o}t{\'e}, M.-A.}, \bibinfo{author}{Houde, J.-C.}, \bibinfo{author}{Descoteaux, M.} \& \bibinfo{author}{Maier-Hein, K.}
\newblock \bibinfo{journal}{\bibinfo{title}{{Fiber tractography using machine learning}}}.
\newblock {\emph{\JournalTitle{Neuroimage}}} \textbf{\bibinfo{volume}{158}}, \bibinfo{pages}{417--429} (\bibinfo{year}{2017}).

\bibitem{cai2024convolutional}
\bibinfo{author}{Cai, L.} \emph{et~al.}
\newblock \bibinfo{title}{{Convolutional-recurrent neural networks approximate diffusion tractography from T1-weighted MRI and associated anatomical context}}.
\newblock In \emph{\bibinfo{booktitle}{Medical Imaging with Deep Learning}}, \bibinfo{pages}{1124--1143} (\bibinfo{year}{2024}).

\bibitem{poulin2017learn}
\bibinfo{author}{Poulin, P.} \emph{et~al.}
\newblock \bibinfo{title}{{Learn to track: deep learning for tractography}}.
\newblock In \emph{\bibinfo{booktitle}{MICCAI}}, \bibinfo{pages}{540--547} (\bibinfo{year}{2017}).

\bibitem{benou2019deeptract}
\bibinfo{author}{Benou, I.} \& \bibinfo{author}{Riklin~Raviv, T.}
\newblock \bibinfo{title}{{Deeptract: A probabilistic deep learning framework for white matter fiber tractography}}.
\newblock In \emph{\bibinfo{booktitle}{MICCAI}}, \bibinfo{pages}{626--635} (\bibinfo{year}{2019}).

\bibitem{wegmayr2021entrack}
\bibinfo{author}{Wegmayr, V.} \& \bibinfo{author}{Buhmann, J.}
\newblock \bibinfo{journal}{\bibinfo{title}{{Entrack: Probabilistic spherical regression with entropy regularization for fiber tractography}}}.
\newblock {\emph{\JournalTitle{International Journal of Computer Vision}}} \textbf{\bibinfo{volume}{129}}, \bibinfo{pages}{656--680} (\bibinfo{year}{2021}).

\bibitem{theberge2021track}
\bibinfo{author}{Th{\'e}berge, A.}, \bibinfo{author}{Desrosiers, C.}, \bibinfo{author}{Descoteaux, M.} \& \bibinfo{author}{Jodoin, P.-M.}
\newblock \bibinfo{journal}{\bibinfo{title}{{Track-to-learn: A general framework for tractography with deep reinforcement learning}}}.
\newblock {\emph{\JournalTitle{Medical Image Analysis}}} \textbf{\bibinfo{volume}{72}}, \bibinfo{pages}{102093} (\bibinfo{year}{2021}).

\bibitem{theberge2024matters}
\bibinfo{author}{Th{\'e}berge, A.}, \bibinfo{author}{Desrosiers, C.}, \bibinfo{author}{Bor{\'e}, A.}, \bibinfo{author}{Descoteaux, M.} \& \bibinfo{author}{Jodoin, P.-M.}
\newblock \bibinfo{journal}{\bibinfo{title}{{What matters in reinforcement learning for tractography}}}.
\newblock {\emph{\JournalTitle{Medical Image Analysis}}} \textbf{\bibinfo{volume}{93}}, \bibinfo{pages}{103085} (\bibinfo{year}{2024}).

\bibitem{janner2022planning}
\bibinfo{author}{Janner, M.}, \bibinfo{author}{Du, Y.}, \bibinfo{author}{Tenenbaum, J.} \& \bibinfo{author}{Levine, S.}
\newblock \bibinfo{title}{{Planning with Diffusion for Flexible Behavior Synthesis}}.
\newblock In \emph{\bibinfo{booktitle}{ICML}}, \bibinfo{pages}{9902--9915} (\bibinfo{year}{2022}).

\bibitem{gu2022stochastic}
\bibinfo{author}{Gu, T.}, \bibinfo{author}{Chen, G.}, \bibinfo{author}{Li, J.} \emph{et~al.}
\newblock \bibinfo{title}{{Stochastic trajectory prediction via motion indeterminacy diffusion}}.
\newblock In \emph{\bibinfo{booktitle}{CVPR}}, \bibinfo{pages}{17113--17122} (\bibinfo{year}{2022}).

\bibitem{lv2024diffmot}
\bibinfo{author}{Lv, W.} \emph{et~al.}
\newblock \bibinfo{title}{{Diffmot: A real-time diffusion-based multiple object tracker with non-linear prediction}}.
\newblock In \emph{\bibinfo{booktitle}{CVPR}}, \bibinfo{pages}{19321--19330} (\bibinfo{year}{2024}).

\bibitem{luo2024diffusiontrack}
\bibinfo{author}{Luo, R.} \emph{et~al.}
\newblock \bibinfo{title}{{Diffusiontrack: Diffusion model for multi-object tracking}}.
\newblock In \emph{\bibinfo{booktitle}{AAAI}}, vol.~\bibinfo{volume}{38}, \bibinfo{pages}{3991--3999} (\bibinfo{year}{2024}).

\bibitem{ho2020denoising}
\bibinfo{author}{Ho, J.}, \bibinfo{author}{Jain, A.} \& \bibinfo{author}{Abbeel, P.}
\newblock \bibinfo{journal}{\bibinfo{title}{{Denoising diffusion probabilistic models}}}.
\newblock {\emph{\JournalTitle{Advances in neural information processing systems}}} \textbf{\bibinfo{volume}{33}}, \bibinfo{pages}{6840--6851} (\bibinfo{year}{2020}).

\bibitem{wegmayr2018data}
\bibinfo{author}{Wegmayr, V.}, \bibinfo{author}{Giuliari, G.}, \bibinfo{author}{Holdener, S.} \& \bibinfo{author}{Buhmann, J.}
\newblock \bibinfo{title}{{Data-driven fiber tractography with neural networks}}.
\newblock In \emph{\bibinfo{booktitle}{ISBI}}, \bibinfo{pages}{1030--1033} (\bibinfo{year}{2018}).

\bibitem{mirzaalian2016inter}
\bibinfo{author}{Mirzaalian, H.}, \bibinfo{author}{Ning, L.} \emph{et~al.}
\newblock \bibinfo{journal}{\bibinfo{title}{{Inter-site and inter-scanner diffusion MRI data harmonization}}}.
\newblock {\emph{\JournalTitle{NeuroImage}}} \textbf{\bibinfo{volume}{135}}, \bibinfo{pages}{311--323} (\bibinfo{year}{2016}).

\bibitem{zhu2024diffusion}
\bibinfo{author}{Zhu, X.}, \bibinfo{author}{Zhang, W.}, \bibinfo{author}{Li, Y.}, \bibinfo{author}{O’Donnell, L.} \& \bibinfo{author}{Zhang, F.}
\newblock \bibinfo{title}{{When diffusion MRI meets diffusion model: A novel deep generative model for diffusion MRI generation}}.
\newblock In \emph{\bibinfo{booktitle}{MICCAI}}, \bibinfo{pages}{530--540} (\bibinfo{year}{2024}).

\bibitem{chi2023diffusion}
\bibinfo{author}{Chi, C.} \emph{et~al.}
\newblock \bibinfo{journal}{\bibinfo{title}{{Diffusion policy: Visuomotor policy learning via action diffusion}}}.
\newblock {\emph{\JournalTitle{The International Journal of Robotics Research}}} \bibinfo{pages}{02783649241273668} (\bibinfo{year}{2023}).

\bibitem{perez2018film}
\bibinfo{author}{Perez, E.}, \bibinfo{author}{Strub, F.}, \bibinfo{author}{De~Vries, H.}, \bibinfo{author}{Dumoulin, V.} \& \bibinfo{author}{Courville, A.}
\newblock \bibinfo{title}{{Film: Visual reasoning with a general conditioning layer}}.
\newblock In \emph{\bibinfo{booktitle}{AAAI}}, vol.~\bibinfo{volume}{32} (\bibinfo{year}{2018}).

\bibitem{theaud2020tractoflow}
\bibinfo{author}{Theaud, G.} \emph{et~al.}
\newblock \bibinfo{journal}{\bibinfo{title}{{TractoFlow: A robust, efficient and reproducible diffusion MRI pipeline leveraging Nextflow \& Singularity}}}.
\newblock {\emph{\JournalTitle{Neuroimage}}} \textbf{\bibinfo{volume}{218}}, \bibinfo{pages}{116889} (\bibinfo{year}{2020}).

\bibitem{cetin2024harmonized}
\bibinfo{author}{Cetin-Karayumak, S.}, \bibinfo{author}{Zhang, F.} \emph{et~al.}
\newblock \bibinfo{journal}{\bibinfo{title}{{Harmonized diffusion MRI data and white matter measures from the Adolescent Brain Cognitive Development Study}}}.
\newblock {\emph{\JournalTitle{Scientific Data}}} \textbf{\bibinfo{volume}{11}}, \bibinfo{pages}{249} (\bibinfo{year}{2024}).

\bibitem{van2013wu}
\bibinfo{author}{Van~Essen, D.}, \bibinfo{author}{Smith, S.} \emph{et~al.}
\newblock \bibinfo{journal}{\bibinfo{title}{{The WU-Minn human connectome project: an overview}}}.
\newblock {\emph{\JournalTitle{Neuroimage}}} \textbf{\bibinfo{volume}{80}}, \bibinfo{pages}{62--79} (\bibinfo{year}{2013}).

\bibitem{maier2017challenge}
\bibinfo{author}{Maier-Hein, K.}, \bibinfo{author}{Neher, P.}, \bibinfo{author}{Houde, J.-C.}, \bibinfo{author}{C{\^o}t{\'e}, M.-A.} \emph{et~al.}
\newblock \bibinfo{journal}{\bibinfo{title}{{The challenge of mapping the human connectome based on diffusion tractography}}}.
\newblock {\emph{\JournalTitle{Nature communications}}} \textbf{\bibinfo{volume}{8}}, \bibinfo{pages}{1349} (\bibinfo{year}{2017}).

\bibitem{poulin2022tractoinferno}
\bibinfo{author}{Poulin, P.}, \bibinfo{author}{Theaud, G.}, \bibinfo{author}{Rheault, F.}, \bibinfo{author}{St-Onge, E.} \emph{et~al.}
\newblock \bibinfo{journal}{\bibinfo{title}{{TractoInferno-A large-scale, open-source, multi-site database for machine learning dMRI tractography}}}.
\newblock {\emph{\JournalTitle{Scientific Data}}} \textbf{\bibinfo{volume}{9}}, \bibinfo{pages}{725} (\bibinfo{year}{2022}).

\bibitem{marek2011parkinson}
\bibinfo{author}{Marek, K.} \emph{et~al.}
\newblock \bibinfo{journal}{\bibinfo{title}{{The Parkinson progression marker initiative (PPMI)}}}.
\newblock {\emph{\JournalTitle{Progress in neurobiology}}} \textbf{\bibinfo{volume}{95}}, \bibinfo{pages}{629--635} (\bibinfo{year}{2011}).

\bibitem{di2017enhancing}
\bibinfo{author}{Di~Martino, A.} \emph{et~al.}
\newblock \bibinfo{journal}{\bibinfo{title}{{Enhancing studies of the connectome in autism using the autism brain imaging data exchange II}}}.
\newblock {\emph{\JournalTitle{Scientific data}}} \textbf{\bibinfo{volume}{4}}, \bibinfo{pages}{1--15} (\bibinfo{year}{2017}).

\bibitem{poldrack2016phenome}
\bibinfo{author}{Poldrack, R.}, \bibinfo{author}{Congdon, E.} \emph{et~al.}
\newblock \bibinfo{journal}{\bibinfo{title}{{A phenome-wide examination of neural and cognitive function}}}.
\newblock {\emph{\JournalTitle{Scientific data}}} \textbf{\bibinfo{volume}{3}}, \bibinfo{pages}{1--12} (\bibinfo{year}{2016}).

\bibitem{garyfallidis2014dipy}
\bibinfo{author}{Garyfallidis, E.}, \bibinfo{author}{Brett, M.} \emph{et~al.}
\newblock \bibinfo{journal}{\bibinfo{title}{{Dipy, a library for the analysis of diffusion MRI data}}}.
\newblock {\emph{\JournalTitle{Frontiers in neuroinformatics}}} \textbf{\bibinfo{volume}{8}}, \bibinfo{pages}{8} (\bibinfo{year}{2014}).

\bibitem{tournier2010improved}
\bibinfo{author}{Tournier, J.~D.}, \bibinfo{author}{Calamante, F.}, \bibinfo{author}{Connelly, A.} \emph{et~al.}
\newblock \bibinfo{title}{{Improved probabilistic streamlines tractography by 2nd order integration over fibre orientation distributions}}.
\newblock In \emph{\bibinfo{booktitle}{ISMRM}}, vol. \bibinfo{volume}{1670}, \bibinfo{pages}{2010} (\bibinfo{year}{2010}).

\bibitem{wasserthal2018tractseg}
\bibinfo{author}{Wasserthal, J.}, \bibinfo{author}{Neher, P.} \& \bibinfo{author}{Maier-Hein, K.}
\newblock \bibinfo{journal}{\bibinfo{title}{{TractSeg-Fast and accurate white matter tract segmentation}}}.
\newblock {\emph{\JournalTitle{Neuroimage}}} \textbf{\bibinfo{volume}{183}}, \bibinfo{pages}{239--253} (\bibinfo{year}{2018}).

\bibitem{yang2025deep}
\bibinfo{author}{Yang, Y.} \emph{et~al.}
\newblock \bibinfo{journal}{\bibinfo{title}{{Deep learning-based diffusion MRI tractography: Integrating spatial and anatomical information}}}.
\newblock {\emph{\JournalTitle{NeuroImage}}} \textbf{\bibinfo{volume}{317}}, \bibinfo{pages}{121314} (\bibinfo{year}{2025}).

\bibitem{zhang2018anatomically}
\bibinfo{author}{Zhang, F.}, \bibinfo{author}{Wu, Y.}, \bibinfo{author}{Norton, I.} \emph{et~al.}
\newblock \bibinfo{journal}{\bibinfo{title}{{An anatomically curated fiber clustering white matter atlas for consistent white matter tract parcellation across the lifespan}}}.
\newblock {\emph{\JournalTitle{Neuroimage}}} \textbf{\bibinfo{volume}{179}}, \bibinfo{pages}{429--447} (\bibinfo{year}{2018}).

\bibitem{zhang2020deep}
\bibinfo{author}{Zhang, F.} \emph{et~al.}
\newblock \bibinfo{journal}{\bibinfo{title}{{Deep white matter analysis (DeepWMA): Fast and consistent tractography segmentation}}}.
\newblock {\emph{\JournalTitle{Medical image analysis}}} \textbf{\bibinfo{volume}{65}}, \bibinfo{pages}{101761} (\bibinfo{year}{2020}).

\bibitem{xue2023superficial}
\bibinfo{author}{Xue, T.}, \bibinfo{author}{Zhang, F.} \emph{et~al.}
\newblock \bibinfo{journal}{\bibinfo{title}{{Superficial white matter analysis: An efficient point-cloud-based deep learning framework with supervised contrastive learning for consistent tractography parcellation across populations and dMRI acquisitions}}}.
\newblock {\emph{\JournalTitle{Medical image analysis}}} \textbf{\bibinfo{volume}{85}}, \bibinfo{pages}{102759} (\bibinfo{year}{2023}).

\bibitem{farquharson2013white}
\bibinfo{author}{Farquharson, S.}, \bibinfo{author}{Tournier, J.-D.}, \bibinfo{author}{Calamante, F.}, \bibinfo{author}{Fabinyi, G.} \emph{et~al.}
\newblock \bibinfo{journal}{\bibinfo{title}{{White matter fiber tractography: why we need to move beyond DTI}}}.
\newblock {\emph{\JournalTitle{Journal of neurosurgery}}} \textbf{\bibinfo{volume}{118}}, \bibinfo{pages}{1367--1377} (\bibinfo{year}{2013}).

\bibitem{vos2013multi}
\bibinfo{author}{Vos, S.~B.}, \bibinfo{author}{Viergever, M.~A.} \& \bibinfo{author}{Leemans, A.}
\newblock \bibinfo{journal}{\bibinfo{title}{{Multi-fiber tractography visualizations for diffusion MRI data}}}.
\newblock {\emph{\JournalTitle{PloS one}}} \textbf{\bibinfo{volume}{8}}, \bibinfo{pages}{e81453} (\bibinfo{year}{2013}).

\bibitem{li2024diffusion}
\bibinfo{author}{Li, Y.} \emph{et~al.}
\newblock \bibinfo{journal}{\bibinfo{title}{{A diffusion MRI tractography atlas for concurrent white matter mapping across Eastern and Western populations}}}.
\newblock {\emph{\JournalTitle{Scientific Data}}} \textbf{\bibinfo{volume}{11}}, \bibinfo{pages}{787} (\bibinfo{year}{2024}).

\bibitem{cousineau2017test}
\bibinfo{author}{Cousineau, M.}, \bibinfo{author}{Jodoin, P.-M.}, \bibinfo{author}{Garyfallidis, E.} \emph{et~al.}
\newblock \bibinfo{journal}{\bibinfo{title}{{A test-retest study on Parkinson's PPMI dataset yields statistically significant white matter fascicles}}}.
\newblock {\emph{\JournalTitle{NeuroImage: Clinical}}} \textbf{\bibinfo{volume}{16}}, \bibinfo{pages}{222--233} (\bibinfo{year}{2017}).

\bibitem{zhang2019test}
\bibinfo{author}{Zhang, F.} \emph{et~al.}
\newblock \bibinfo{journal}{\bibinfo{title}{{Test--retest reproducibility of white matter parcellation using diffusion MRI tractography fiber clustering}}}.
\newblock {\emph{\JournalTitle{Human brain mapping}}} \textbf{\bibinfo{volume}{40}}, \bibinfo{pages}{3041--3057} (\bibinfo{year}{2019}).

\bibitem{cote2013tractometer}
\bibinfo{author}{C{\^o}t{\'e}, M.-A.} \emph{et~al.}
\newblock \bibinfo{journal}{\bibinfo{title}{{Tractometer: towards validation of tractography pipelines}}}.
\newblock {\emph{\JournalTitle{Medical image analysis}}} \textbf{\bibinfo{volume}{17}}, \bibinfo{pages}{844--857} (\bibinfo{year}{2013}).

\bibitem{garyfallidis2018recognition}
\bibinfo{author}{Garyfallidis, E.}, \bibinfo{author}{C{\^o}t{\'e}, M.-A.}, \bibinfo{author}{Rheault, F.} \emph{et~al.}
\newblock \bibinfo{journal}{\bibinfo{title}{{Recognition of white matter bundles using local and global streamline-based registration and clustering}}}.
\newblock {\emph{\JournalTitle{NeuroImage}}} \textbf{\bibinfo{volume}{170}}, \bibinfo{pages}{283--295} (\bibinfo{year}{2018}).

\bibitem{zhang2021deep}
\bibinfo{author}{Zhang, F.}, \bibinfo{author}{Wells, W.} \& \bibinfo{author}{O’Donnell, L.}
\newblock \bibinfo{journal}{\bibinfo{title}{{Deep diffusion MRI registration (DDMReg): a deep learning method for diffusion MRI registration}}}.
\newblock {\emph{\JournalTitle{IEEE transactions on medical imaging}}} \textbf{\bibinfo{volume}{41}}, \bibinfo{pages}{1454--1467} (\bibinfo{year}{2021}).

\bibitem{golby2011interactive}
\bibinfo{author}{Golby, A.} \emph{et~al.}
\newblock \bibinfo{journal}{\bibinfo{title}{{Interactive diffusion tensor tractography visualization for neurosurgical planning}}}.
\newblock {\emph{\JournalTitle{Neurosurgery}}} \textbf{\bibinfo{volume}{68}}, \bibinfo{pages}{496--505} (\bibinfo{year}{2011}).

\bibitem{leclercq2011diffusion}
\bibinfo{author}{Leclercq, D.}, \bibinfo{author}{Delmaire, C.}, \bibinfo{author}{de~Champfleur, N.}, \bibinfo{author}{Chiras, J.} \& \bibinfo{author}{Leh{\'e}ricy, S.}
\newblock \bibinfo{journal}{\bibinfo{title}{{Diffusion tractography: methods, validation and applications in patients with neurosurgical lesions}}}.
\newblock {\emph{\JournalTitle{Neurosurgery Clinics}}} \textbf{\bibinfo{volume}{22}}, \bibinfo{pages}{253--268} (\bibinfo{year}{2011}).

\bibitem{yamada2009mr}
\bibinfo{author}{Yamada, K.}, \bibinfo{author}{Sakai, K.}, \bibinfo{author}{Akazawa, K.}, \bibinfo{author}{Yuen, S.} \& \bibinfo{author}{Nishimura, T.}
\newblock \bibinfo{journal}{\bibinfo{title}{{MR tractography: a review of its clinical applications}}}.
\newblock {\emph{\JournalTitle{Magnetic resonance in medical sciences}}} \textbf{\bibinfo{volume}{8}}, \bibinfo{pages}{165--174} (\bibinfo{year}{2009}).

\bibitem{yu2005diffusion}
\bibinfo{author}{Yu, C.}, \bibinfo{author}{Li, K.}, \bibinfo{author}{Xuan, Y.}, \bibinfo{author}{Ji, X.} \& \bibinfo{author}{Qin, W.}
\newblock \bibinfo{journal}{\bibinfo{title}{{Diffusion tensor tractography in patients with cerebral tumors: a helpful technique for neurosurgical planning and postoperative assessment}}}.
\newblock {\emph{\JournalTitle{European journal of radiology}}} \textbf{\bibinfo{volume}{56}}, \bibinfo{pages}{197--204} (\bibinfo{year}{2005}).

\bibitem{essayed2017white}
\bibinfo{author}{Essayed, W.} \emph{et~al.}
\newblock \bibinfo{journal}{\bibinfo{title}{{White matter tractography for neurosurgical planning: A topography-based review of the current state of the art}}}.
\newblock {\emph{\JournalTitle{NeuroImage: Clinical}}} \textbf{\bibinfo{volume}{15}}, \bibinfo{pages}{659--672} (\bibinfo{year}{2017}).

\bibitem{vaswani2017attention}
\bibinfo{author}{Vaswani, A.}, \bibinfo{author}{Shazeer, N.}, \bibinfo{author}{Parmar, N.} \emph{et~al.}
\newblock \bibinfo{journal}{\bibinfo{title}{{Attention is all you need}}}.
\newblock {\emph{\JournalTitle{Advances in neural information processing systems}}} \textbf{\bibinfo{volume}{30}} (\bibinfo{year}{2017}).

\bibitem{gu2023mamba}
\bibinfo{author}{Gu, A.} \& \bibinfo{author}{Dao, T.}
\newblock \bibinfo{journal}{\bibinfo{title}{{Mamba: Linear-time sequence modeling with selective state spaces}}}.
\newblock {\emph{\JournalTitle{arXiv preprint arXiv:2312.00752}}}  (\bibinfo{year}{2023}).

\end{thebibliography}

\end{document}